\documentclass{article}
\usepackage[nonatbib,final]{neurips_2020}

\usepackage[utf8]{inputenc} % allow utf-8 input
\usepackage[T1]{fontenc}    % use 8-bit T1 fonts
\usepackage{url}            % simple URL typesetting
\usepackage{commath}
\usepackage{booktabs}       % professional-quality tables
\usepackage{amsfonts}       % blackboard math symbols
\usepackage{nicefrac}       % compact symbols for 1/2, etc.
\usepackage{microtype}      % microtypography
\usepackage{iac_pkg}
\usepackage{xspace}
\usepackage{graphicx}
\usepackage{subfig}
\usepackage{xcolor, colortbl}

\definecolor{citecolor}{RGB}{34,139,34}
\definecolor{tablehighlight}{gray}{0.85}
\usepackage{amsmath,amssymb,amsfonts,flushend,tabularx}
\usepackage[pagebackref=true,breaklinks=true,letterpaper=true,citecolor=citecolor,colorlinks,bookmarks=false]{hyperref}
\usepackage[capitalise]{cleveref}

\DeclareMathOperator*{\argmin}{argmin}

\begin{document}

\title{Evaluating the Robustness of Geometry-Aware Instance-Reweighted Adversarial Training}

\author{
  Dorjan Hitaj\thanks{Authors contributed equally.} \\
  Department of Computer Science\\
  Sapienza University of Rome\\
  \texttt{hitaj.d@di.uniroma1.it} \\
   \And
  Giulio Pagnotta\footnotemark[1] \\
  Department of Computer Science\\
  Sapienza University of Rome\\
  \texttt{pagnotta@di.uniroma1.it} \\
  \And
  Iacopo Masi \\
  Department of Computer Science\\
  Sapienza University of Rome\\
  \texttt{masi@di.uniroma1.it} \\
   \And
  Luigi V. Mancini \\
  Department of Computer Science\\
  Sapienza University of Rome\\
  \texttt{mancini@di.uniroma1.it} \\

}

% \author{
% }
\maketitle

\begin{abstract}
In this technical report, we evaluate the adversarial robustness of a very recent method called ``Geometry-aware Instance-reweighted Adversarial Training''~\cite{zhang2020geometry}. GAIRAT reports state-of-the-art results on defenses to adversarial attacks on the CIFAR-10 dataset. In fact, we find that a network trained with this method, while showing an improvement over regular adversarial training (AT), is \emph{biasing} the model towards certain samples by re-scaling the loss. Indeed, this leads the model to be susceptible to attacks that scale the logits.
The original model shows an accuracy of 59\% under AutoAttack – when trained with additional data with pseudo-labels. We provide an analysis that shows the opposite. In particular, we craft a PGD attack multiplying the logits by a positive scalar that decreases the GAIRAT accuracy from from 55\% to 44\%, when trained solely on CIFAR-10.
In this report, we rigorously evaluate the model and provide insights into the reasons behind the vulnerability of GAIRAT to this adversarial attack.
The code to reproduce our evaluation is made available at \href{https://github.com/giuxhub/GAIRAT-LSA}{https://github.com/giuxhub/GAIRAT-LSA}
\end{abstract}
\section{Introduction}

Geometry-Aware Instance-Reweighted Adversarial Training (GAIRAT)~\cite{zhang2020geometry} proposes a method to adversarially train a deep neural network (DNN) by imposing the notion of \textit{importance} on adversarial data points used in the training procedure. This \textit{importance} is measured by approximating the distance to the class boundary of natural data points.
The notion behind GAIRAT is that a natural data point that is closer to the class boundary is less robust since it is easier to flip its label; vice versa, if the natural datum is farther from the boundary then is referred to as ``guarded data'', or else, harder to attack. This intuition guides the method to treat training points differently during adversarial risk minimization. The adversarial training samples should be assigned with different weights in the loss: points closer to the class boundary should be penalized more (have higher weights in the loss), otherwise points far away from the boundary should be considered less in the training (smaller weights).
Based on this insight, GAIRAT~\cite{zhang2020geometry} approximates the geometric distance of a natural data point by using the least number of iterations \textbf{$\kappa$} that projected gradient descent (PGD)~\cite{PGDMadry} requires to generate a misclassified adversarial instance from that natural data point. GAIRAT explicitly assigns instance-dependent weight to the loss of the adversarial variant of natural data using a function $\omega(\mbf{x},y,\kappa)$ where $\mbf{x}$ is the training sample, $y$ is the label and \textbf{$\kappa$} indicates the ``distance'' from the boundary.

However, being a very recent approach, to the best of our knowledge, the vulnerabilities and potential weaknesses of GAIRAT have not been previously studied thoroughly. In fact, in this report we show that, even though being able to maintain a high performance on natural data, when it comes to adversarial robustness, GAIRAT is vulnerable towards a logit scaling attack.
\newline

The rest of this technical report is organized as follows: \cref{sec:Preliminaries} gives a brief description of the notion of adversarial training as used by \cite{PGDMadry} and \cite{zhang20FAT} and how GAIRAT builds on top of that to attempt to enforce robustness in the model. In~\cref{sec:scale_logit_attack} we present the logit scaling attack following~\cite{croce2020reliable}. \cref{sec:expt} provides an experimental evaluation and analysis on the effectiveness of our attack and ~\cref{sec:conclusion} concludes this report.

\section{Preliminaries and Overview of the Methods}\label{sec:Preliminaries}

\subsection{Adversarial Training}\label{sec:ATandPGD}
Adversarial examples are inputs to machine learning models that an attacker has intentionally crafted to cause the model to make a mistake. Intuitively, to make the ML models robust to these adversarial examples, they are trained also on these adversarial examples a.k.a. \textit{adversarial training}. 
When it comes to performing \textit{adversarial training} one of the key issues is to decide on which of these adversarial examples a ML model should be trained so that it can become somehow robust towards these malicious data points.
Below we describe two techniques for performing adversarial training, namely \textit{adversarial training} by Madry \etal~\cite{PGDMadry} and \textit{friendly adversarial training} by Zhang \etal~\cite{zhang20FAT}. 
Moreover, we describe how these two techniques are exploited to build another adversarial training technique i.e. GAIRAT~\cite{zhang2020geometry}.

Let \( (\mathcal{X}, d_{\infty}) \) denote the input feature space \(\mathcal{X}\) with the distance metric \(d_{\inf}(x,x^{'}) = \norm{x - x^{'}}_{\infty}\),  and \(B_{\epsilon}[x] = \{x^{'} \in \mathcal{X} ~|~d_{\inf}(x,x^{'}) \leq \epsilon  \}\) be the closed ball of radius \(\epsilon > 0\) centered at \(x\) in \(\mathcal{X}\).
\(S = \{(x_{i}, y_{i})\}_{i=1}^n\) represents the dataset, where \(x_{i} \in \mathcal{X}\) and \(y_{i} \in \mathcal{Y} = \{0,1,\dots,\mathcal{C}-1\}\) indicate the high-dimensional input point and the corresponding label respectively.

\minisection{Adversarial Training (AT)} Given that, the objective function of \textit{standard adversarial training (AT)}~\cite{PGDMadry} is:

\begin{equation}\label{eq:objective_func_Madry}
\min_{f_{\theta} \in \mathcal{F}} \frac{1}{n} \sum_{i=1}^{n} l\Big(f_{\theta}(\Tilde{x}, y_{i})\Big) \qquad \Tilde{x}_{i} = {\arg \max}_{\Tilde{x}\in \mathcal{B}_{\epsilon}[x_{i}]} l\Big(f_{\theta}(\Tilde{x}), y_{i}\Big)
\end{equation}

where \(\Tilde{x}\) is the most adversarial data within the \(\epsilon\)-ball centered at \(\Tilde{x}\), 
\(f_{\theta}(\cdot) : \mathcal{X} \rightarrow \mathbb{R}^{\mathcal{C}}\) is a score function, and the loss function \(l: \mathbb{R}^{\mathcal{C}} \times \mathcal{Y} \rightarrow \mathbb{R} \) is a composition of a base loss \(l_{B}: \triangle^{\mathcal{C}-1} \times \mathcal{Y} \rightarrow \mathbb{R} \) and an inverse link function \(l_{L} :  \mathbb{R}^{\mathcal{C}} \rightarrow \triangle^{\mathcal{C}-1} \), in which \(\triangle^{\mathcal{C}-1} \) is the corresponding probability simplex. Differently put, \(l(f_{\theta}(\cdot), y) = l_{B}(l_{L}(f_{\theta}(\cdot)),y) \).
AT~\cite{PGDMadry} uses the most adversarial data generated according to right most part of \cref{{eq:objective_func_Madry}} to update the current model.

\minisection{Friendly Adversarial Training (FAT)}  On other hand, the objective function of \textit{friendly adversarial training} (FAT)~\cite{zhang20FAT} is

\begin{equation}\label{eq:fat}
    \Tilde{x} = \argmin_{\Tilde{x} \in \mathcal{B}_{\epsilon}[x_{i}]}l(f_{\theta}(\Tilde{x}, y_{i}))~\text{s.t. l}~(f_{\theta}(\Tilde{x}, y_{i})) - {\min}_{y\in\mathcal{Y}} l(f(\Tilde{x}), y) \geq p 
\end{equation}

The objective function of AT and FAT implies the optimization of adversarially robust networks, by having one step that generates the adversarial data and one step that minimizes the loss on the generated adversarial data w.r.t. the model parameters \(\theta\).

The projected gradient descent method (PGD)~\cite{PGDMadry} is the most common approximation method for searching adversarial data. Given a starting point \( x^{(0)} \in \mathcal(X)\) and step size \(\lambda > 0 \), PGD works as follows:

\begin{equation}\label{eq:pgd}
    x^{(t+1)} = {\prod}_{\mathcal{B}[x^{(0)}]}\Big(x^{(t)} + \lambda~\text{sign}(\nabla_{x^{(t)}} l(f_{\theta}(x^{(t)}),y))\Big) , t \in \mathbb{N}
\end{equation}

until a stopping criterion is satisfied. In~\cref{eq:pgd} \textit{l} is the loss function, \(x^{(0)}\) refers to natural data or natural data perturbed by a small Gaussian or uniformly random noise; \textit{y} is the corresponding label for natural data; \(x^{(t)}\) is adversarial data at iteration \textit{t}; and \( {\prod}_{\mathcal{B}[x^{(0)}]}(\cdot)\) is the projection function that projects the adversarial data back into the \(\epsilon\)-ball centered at \(x^{(0)}\) if needed.

\subsection{GAIRAT}\label{sec:gairat_technique}

Let \(\omega(x,y)\) be the \textit{geometry-aware weight assignment function} on the loss of the adversarial variant \(\Tilde{x}\). The inner optimization for generating \(\Tilde{x}\) still follows the right part of ~\cref{eq:objective_func_Madry} or~\cref{eq:fat}. The outer minimization is changed to:
\begin{equation}
   \min_{f_{\theta} \in \mathcal{F}} \frac{1}{n} \sum_{i=1}^{n} \omega{(x_{i}, y_{i})}l{\Big(f_{\theta}(x_{i}),y_{i}\Big)}.
   \label{eq:gairat}
\end{equation}

\cref{eq:gairat} rescales the loss using a function \(\omega{(x_{i}, y_{i})}\). This function is non-increasing w.r.t. the \textit{geometric-distance} $\kappa$. $\kappa$ is a counter in the PGD optimization that counts how many PGD iteration the point keep its original labels and implicitly encodes the distance from data \(x_{i}\) to the decision boundary. The method then normalizes $\omega$ so that \(\omega{(x_{i}, y_{i})} \geq 0\) and \(\frac{1}{n}  \sum_{i=1}^{n} \omega{(x_{i}, y_{i}) = 1}\). 

Thus, \cref{eq:gairat} does not consider those points that are significantly far away from the decision boundary, whereas it sets larger weights for those \textit{x} close to the decision boundary.

Finally GAIRAT employs a \textit{bootstrap period} in the initial part of the training by setting \(\omega{(x_{i}, y_{i})} = 1\), thereby performing regular training, regardless of the \textit{geometric-distance} of input \((x_{i}, y_{i})\). This is motivated for the fact that distance metric is less informative when the classifier is at the beginning with boundaries that are not properly learned.
\section{Logit Scaling Attack}\label{sec:scale_logit_attack}
\minisection{Rationale for the attack} As we have observed in~\cref{sec:gairat_technique}, GAIRAT assigns different weights to different training samples based on their vicinity to the class boundaries. This induces an apparent robustness when the method is tested with vanilla PGD. Training in such way biases the model to ignore some training samples and increasing the emphasis on other points.
Following the observation in~\cite{CarliniW16, croce2020reliable}, we notice that the cross-entropy loss (CE) is not scale-invariant, in other words, the model loss is highly sensitive to scaling in the logits. Despite GAIRAT does not directly change the logits values, it does re-scale the loss using the $\omega$ function. 
\newline
In this report we show that, in the same way Croce and Hein did in~\cite{croce2020reliable}, is possible to decrease the robust accuracy in a non negligible way. Thus we changed the PGD optimization described in~\cref{eq:pgd} so that it incorporate a positive scalar value $\alpha > 0$ that is multiplied to the logits of the GAIRAT model, \emph{before} applying softmax operator that maps the logits on the probability simplex. The $\alpha$ value is found through linear search. Yet, this is sufficient to empirically show that is possible to find points in the input space that decrease the GAIRAT accuracy of about 11\%.
It can be shown that the softmax function can be parametrized by $\alpha$ to span a family of functions that changes the shape of the output distribution, given a fixed value of the logits. In particular, the effect of alpha is that for $0 < \alpha < 1$, the softmax output approaches the uniform distribution, whereas $\alpha \gg 1$ peaks the distribution towards the max value of the logits. By doing so, the method attains stronger gradient on the input bypassing the defense, thereby restoring the gradient flow from the loss to the input image. This is also verified in Fig.2 in~\cite{croce2020reliable} where increasing the scale $\alpha$ decreases the percentage of zeros in the gradient therefore prevent gradient masking. In~\cref{sec:expt} we offer a version of PGD$_{\alpha=10}$ that implements the logit scaling attack.

\minisection{Inference time} Finally we note that $\alpha$ is just used when crafting the attack. Once the adversarial images are found, $\alpha$ does not influence the testing, and even if it did, the argmax operation for selecting the predicted class label is scale invariant.

\section{Experimental Validation and Analysis}\label{sec:expt}
We perform a simple evaluation of the GAIRAT model following the training setting reported in the paper and using the publicly available source code released in \href{https://github.com/zjfheart/Geometry-aware-Instance-reweighted-Adversarial-Training}{github.com/zjfheart/Geometry-aware-Instance-reweighted-Adversarial-Training}.
To marginalize out the effect of additional data~\cite{NEURIPS2019_32e0bd14} used for training, we have re-trained the GAIRAT model only on CIFAR-10 training images.

\minisection{Threat Model and Experimental Settings} To evaluate the GAIRAT approach, the paper~\cite{zhang2020geometry} tests it in the white-box setting, with the space of the attacker being a $\ell_{\infty}$ ball of size $0.031$ around each natural image.
The evaluation reports the regular test accuracy along with the robust accuracy on adversarial data generated using PGD-20 and PGD+. 
PGD-20 follows the same setting used by~\cite{pmlr-v97-wang19i}, where the value $20$ indicates the iteration of PGD. The model also undergoes a tougher attack, which is the PGD+ configuration specified in~\cite{NEURIPS2019_32e0bd14}.

The adversarial bounds have the same \(\epsilon_{test} = 0.031\). For PGD-20, the number of iteration in PGD is 20, step size \(\lambda = \epsilon_{test}/4\), one random start that samples the perturbation from an uniform distribution in \(([-\epsilon_{test}, +\epsilon_{test}])\) added to the natural data before the PGD perturbation. 

Following~\cite{NEURIPS2019_32e0bd14}, PGD+ is configured with 40 iterations, with a step size of  \(\lambda = 0.01\) and 5 random restarts for each natural test data. Besides a larger number of iterations and more stochasticity in the initialization, PGD+ is a tougher attack since during the attack optimization the point is optimized for $200$ iterations and an image is considered correct in case does not change the correct label across \emph{all} the $200$ iterations; that is, a simple misclassification of the label in the optimization marks the image as incorrect.

%% ---------------------------------------------------------------------------%
\begin{table}[h]
\centering
\begin{tabular}{l ccc cc}
\toprule
 & \multicolumn{3}{c}{Reported} & \multicolumn{2}{c}{Our evaluation}\\
{Method} & Natural & PGD-20 & PGD+ & PGD$_{\alpha=10}$ & PGD+$_{\alpha=10}$ \\
\cmidrule(r){1-1} \cmidrule(l){2-4} \cmidrule(l){5-6}
AT, Madry \etal ~\cite{PGDMadry}  & 86.62	& 46.73	& \tbf{46.08}	& 52.61 & 49.96 \\
FAT, Zhang \etal ~\cite{zhang20FAT} & 88.18	& 46.79	& \tbf{45.80}  & 68.63 & 69.30 \\
GAIRAT, Zhang \etal ~\cite{zhang2020geometry}  &  85.49	& 53.76	& 50.32 & 43.89 & \tbf{43.27} \\
\cmidrule(r){1-1}  \cmidrule(l){2-4} \cmidrule(l){5-6}
\end{tabular}

\caption{\tbf{Robust accuracy on CIFAR-10.} Test accuracy of WRN-32-10 on CIFAR-10 dataset when the model are trained \emph{solely} on CIFAR-10 training split. Bold numbers indicate the worst testing accuracy for each row.}
\vspace{-8pt}
\label{tab:accuracy_measures}
\end{table}
%% ---------------------------------------------------------------------------%
%% ---------------------------------------------------------------------------%

%% ---------------------------------------------------------------------------%
\begin{figure*}[!h]
    \subfloat[PGD-20 performance\label{fig:pgd_20}]{
        \includegraphics[width=.475\textwidth]{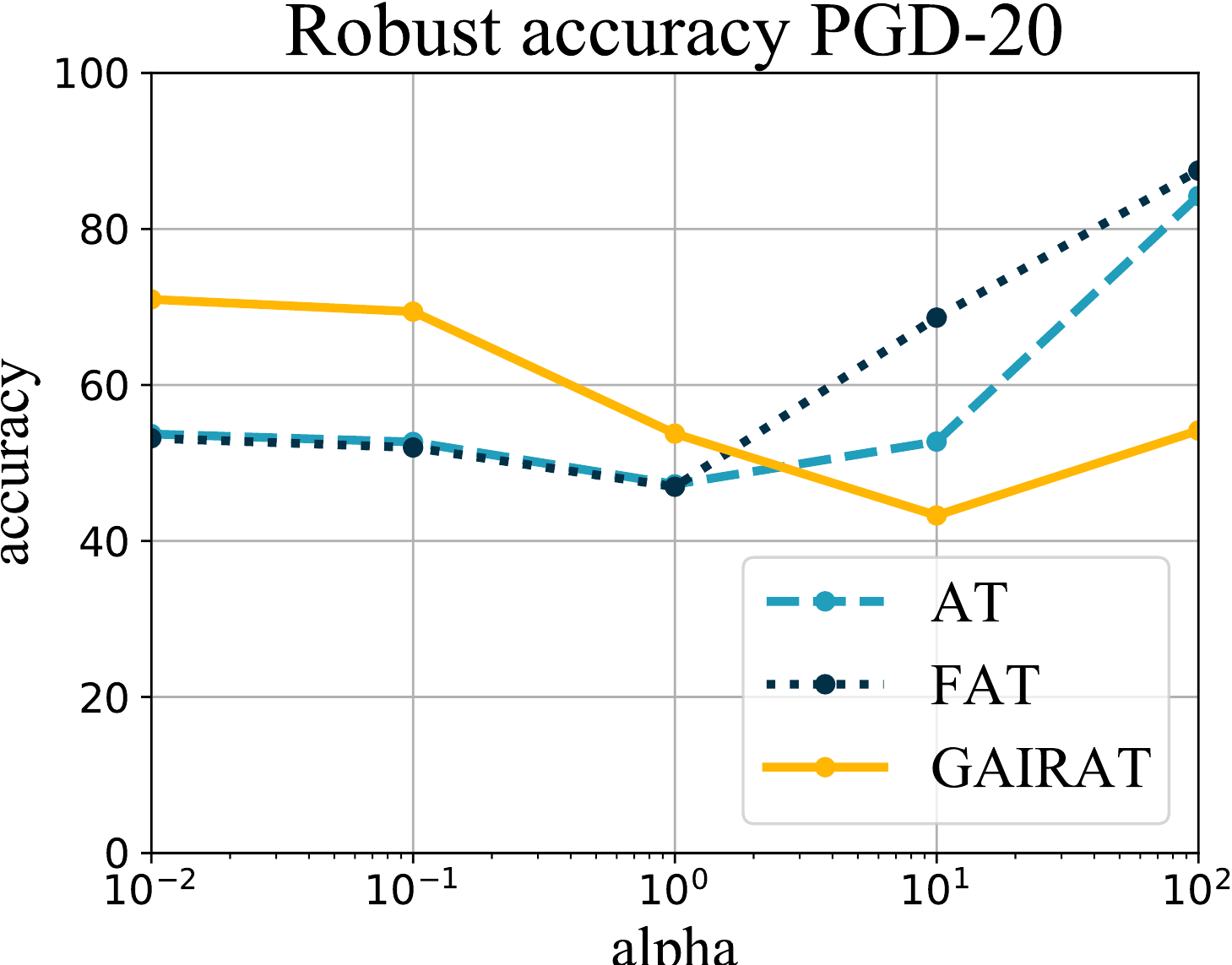}
    }\quad
    \subfloat[PGD+ performance \label{fig:pgd_plus}]{
        \includegraphics[width=.475\textwidth]{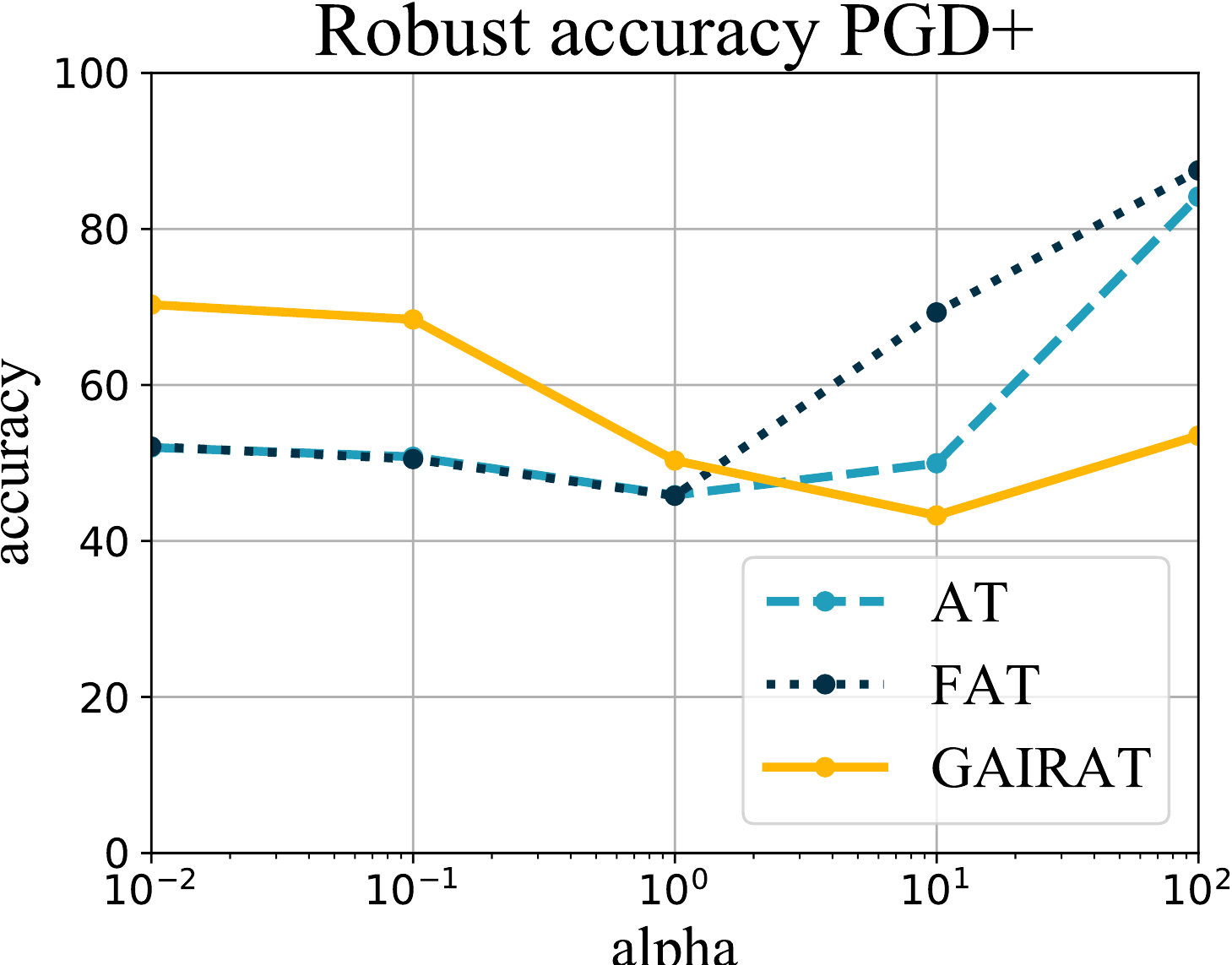}
    }
    \caption{Performance of AT~\cite{PGDMadry}, FAT~\cite{zhang20FAT} and GAIRAT~\cite{zhang2020geometry} on PGD-20 and PGD+ with various $\alpha$ values.}
    \label{fig:pgd_results}
\end{figure*}
%% ---------------------------------------------------------------------------%
%% ---------------------------------------------------------------------------%

\minisection{Evaluation and Analysis} We report our evaluation in \cref{tab:accuracy_measures} under the threat model and settings mentioned above. The table reports the accuracy on CIFAR-10 using the public available model of Madry \etal trained with Adversarial Training (AT) along with the model trained with ``Friendly AT'' in FAT. Finally we report the accuracy of the GAIRAT model~\cite{zhang2020geometry}. The table offers the natural accuracy and the results from the respective papers under PGD-20 and PGD+. On the most right part, we report our evaluation with the logit scaling attack using $\alpha=10$.
We note how AT and FAT are still sensitive to $\alpha$ since even in their case the loss employed is not scale invariant, thus we can still change the optimization for the input points by tuning the scale $\alpha$. The difference is in the fact that while for AT and FAT the worst testing accuracy coincides with $\alpha=1$---thus increasing $\alpha$ leads to an actually boost in the accuracy since gradients flow less in this case---the outcome of GAIRAT is the opposite.
We found that GAIRAT reduces their accuracy from 53.75\% to 43.89\% when using $\alpha=10$ thereby restoring the gradients in the input and making the model susceptible to PGD once again. In this sense, despite GAIRAT training rationale is interesting and brings fresh ideas to the adversarial defense, it leads to a false sense of security and is more vulnerable than AT. Note, however, that this attack does not bring GAIRAT accuracy to zero leaving some level of defense.
Following~\cite{croce2020reliable}, we report in \cref{fig:pgd_results} the full evaluation in function of the scale $\alpha$ ranging from $10^{-2}$ to $10^2$. It is clear from both plots that while AT and FAT the worst case coincides with $\alpha=1$ which correspond to the classic way of training the method, the model GAIRAT reports its worst accuracy with $\alpha=10$. This observation holds for both the easier PGD-20 attack (\cref{fig:pgd_20}) as well for the more difficult PGD+ (\cref{fig:pgd_plus}).

\minisection{Evaluation with more PGD iterations} We also note that the type of training reported in~\cref{sec:gairat_technique} induces very subtle gradient masking artifacts since it is even insensitive to the increase in the number of iteration of PGD. As the authors have suggested in \footnote{https://openreview.net/forum?id=iAX0l6Cz8ub} the evaluation of PGD-200 with a very fine-grained step size does not scratch much the robust accuracy, giving a false sense of robustness~\cite{athalye2018obfuscated}.

\section{Conclusions}\label{sec:conclusion}
We have shown how biasing the cross-entropy loss in a DNN, when trained for adversarial risk minimization, can induce subtle gradient masking which, in turn, is hard to catch and detect unless the attacker performs an attack that scales the logits, as suggested in~\cite{croce2020reliable}. 
Our evaluation analysis does not yet cover whether the same drop in accuracy is attainable simply by running the ensemble attack proposed by AutoAttack~\cite{croce2020reliable}, however it leads us to believe that GAIRAT model trained only on CIFAR-10 may be susceptible to AutoAttack. In particular, we plan to investigate the impact of the Difference of Logits Ratio Loss (DLR) loss proposed in~\cite{croce2020reliable} to this drop in the accuracy. 
Logit scaling attack is not optimal since it depends on a single scalar value that needs to be sough through linear search. Nevertheless, though we believe that the ideas and rationale behind GAIRAT~\cite{zhang2020geometry} are fresh and interesting, this issue still sends an ``alarm bell'' that adversarial defenses are very fragile and hard to assess.

\bibliographystyle{splncs04}
\bibliography{bibliography}

\end{document}